# Should We Fear Large Language Models?
## -A Structural Analysis of the Human Reasoning System for Elucidating LLM Capabilities and Risks Through the Lens of Heidegger's Philosophy

Jianqiu Zhang
The University of Texas at San Antonio
michelle.zhang@utsa.edu

Abstract

In the rapidly evolving field of Large Language Models (LLMs), there is a critical need to thoroughly analyze their capabilities and risks. Central to our investigation are two novel elements. Firstly, it is the innovative parallels between the statistical patterns of word relationships within LLMs and Martin Heidegger's concepts of "ready-to-hand" and "present-at-hand," which encapsulate the utilitarian and scientific altitudes humans employ in interacting with the world. This comparison lays the groundwork for positioning LLMs as the digital counterpart to the Faculty of Verbal Knowledge, shedding light on their capacity to emulate certain facets of human reasoning. Secondly, a structural analysis of human reasoning, viewed through Heidegger's notion of truth as "unconcealment" is conducted This foundational principle enables us to map out the inputs and outputs of the reasoning system and divide reasoning into four distinct categories. Respective cognitive faculties are delineated, allowing us to place LLMs within the broader schema of human reasoning, thus clarifying their strengths and inherent limitations. Our findings reveal that while LLMs possess the capability for *Direct Explicative Reasoning* and *Pseudo Rational Reasoning*, they fall short in authentic rational reasoning and have no creative reasoning capabilities, due to the current lack of many analogous AI models such as the Faculty of Judgement. The potential and risks of LLMs when they are augmented with other AI technologies are also evaluated. The results indicate that although LLMs have achieved proficiency in some reasoning abilities, the aspiration to match or exceed human intellectual capabilities is yet unattained. This research not only enriches our comprehension of LLMs but also propels forward the discourse on AI's potential and its bounds, paving the way for future explorations into AI's evolving landscape.

Introduction

The trajectory of Large Language Models (LLMs) represents a monumental leap in artificial intelligence. Initially, models like GPT-3 broke new ground with their capacity to produce text that mirrors human writing, thanks to their training on expansive datasets. The advent of Reinforcement Learning from Human Feedback (RLHF) (Christiano et al. 2017) marked a pivotal evolution, enhancing LLMs' ability to tailor their outputs more closely to human preferences. This advancement has been crucial for the development of sophisticated models such as ChatGPT and GPT-4, which have demonstrated remarkable progress in mimicking human-like text comprehension and generation. Particularly, GPT-4 has astounded the AI community by surpassing numerous academic and professional standards, including excelling in bar, SAT, and LSAT exams (Zhong et al. 2023); (Katz et al. 2023), showcasing not just academic excellence but also practical utility, as seen in its application by Casetext's CoCounsel (Schwarcz and Choi 2023) for streamlining legal document analysis.

Moreover, GPT-4 has excelled in specialized fields, from conceptual physics (West 2023) to radiology (Bhayana, Bleakney, and Krishna 2023), and enhanced computational capabilities through integrations with tools like Code Interpreter. Concurrently, Google's PaLM 2(Anil et al. 2023) has advanced the frontier with compute-optimal scaling, enriched dataset mixtures, and architectural enhancements, setting new records in reasoning, coding, multilingualism, and natural language generation.

Specialized models such as Instruct GPT(Ouyang et al. 2022), Google's LaMDA (Thoppilan et al. 2022), and the Megatron-Turing NLG (Smith et al. 2022) have each extended the capabilities of LLMs in unique ways. Instruct GPT has improved AI's user-friendliness by optimizing for instruction-following, while LaMDA has revolutionized conversational AI with its unparalleled dataset size. The Megatron-Turing NLG, with its staggering 530 billion parameters, epitomizes the scale and ambition driving current LLM development.

The exploration into LLMs' abstract reasoning has also opened new vistas for reinforcement learning applications, from robotics to embodied AI (W. Huang et al. 2022; Wang et al. 2024; Song et al. 2022), showcasing LLMs' potential to transcend natural language processing and impact a wide spectrum of AI endeavors. The emergence of multimodal LLMs, capable of processing and generating not just text but also astounding images, audio, and video, heralds a new era of AI applications. OpenAI's DALL·E and the most recent Sora, alongside Google's Imagen and Gemini, exemplify this trend, underscoring the expansive potential of foundational models.

Overall, the evolution of LLMs is characterized by rapid advancements across multiple dimensions, including model size, efficiency, contextual understanding, multimodality, ethical AI development, and application specificity, indicating a broad and deepening impact on the landscape of artificial intelligence.

The burgeoning achievements of LLMs have sparked an intense debate on the dual-edged sword of artificial intelligence advancements. Jeffrey Hinton, a seminal figure dubbed the 'godfather of AI,' has spotlighted existential concerns stemming from the unbridled progression of AI technologies (Anderson 2023). Amplifying these apprehensions, a collective call for a moratorium on the development of sophisticated AI systems, endorsed by over 33,000 tech leaders and academics, highlights critical ethical dilemmas. The letter raises several pressing, unanswered questions: Should we permit machines to disseminate propaganda and falsehoods unchecked? Should we automate even fulfilling jobs, potentially rendering human skills obsolete? Should we allow the evolution of non-human intellects that could eventually surpass and supplant humanity? And, ultimately, are we willing to risk losing control of our own civilization? (FLI staff 2023)

Addressing the ethical dilemmas that arise with AI advancements necessitates a comprehensive analytical approach to compare the capabilities of LLMs with those of human intellect, particularly examining AI's potential to meet or surpass human cognitive capabilities. While human cognition encompasses a wide array of capabilities, reasoning is a pivotal component. Our discussion predominantly revolves around LLMs due to their significant overlap with reasoning processes. However, there's a discernible lack of clarity within academic circles about how LLMs compare with human reasoning systems, and the extent to which LLMs contribute to or replicate these cognitive processes remains underexplored.

Scholarly critiques have shed some light on LLMs' limitations, such as inaccuracies in responses, failure in logical reasoning, and linguistic inconsistencies. (Floridi and Chiriatti 2020; Cobbe et al. 2021; Perez and Ribeiro 2022; Arkoudas 2023b; Floridi 2023); (Ettinger 2020; Traylor, Feiman, and Pavlick 2021), compositionality (Kim and Linzen 2020; Yu and Ettinger 2020) and generalization outside of their training distribution (Glockner, Shwartz, and Goldberg 2018; Jia and Liang 2017; Thomas McCoy, Pavlick, and Linzen 2019).

The core of these limitations is often attributed to neural networks' inability to process abstract symbolic structures, crucial for tasks involving variable binding and logical deduction (Quilty-Dunn, Porot, and Mandelbaum 2022). Furthermore, the reliance on text for training limits LLMs' understanding of prompts related to tangible entities, highlighting a gap in their comprehension of the real world (Bisk et al. 2020; Bender and Koller 2020). While models like OpenAI's latest, Sora, have advanced by incorporating video images alongside text in their training data, they still fall short of achieving a genuine comprehension of the world. This shortfall is evident in their inability to precisely simulate the physics of complex scenes and in a lack of understanding of specific instances of cause and effect (OpenAI 2024).

The current literature, while valuable, lacks a structured methodology to fully conduct the comparative analysis of LLMs and human reasoning. Simplifying the discussion to the encoding of abstract symbolic structures overlooks the multifaceted nature of rational thought and the complexities inherent in human reasoning beyond rational reasoning. Prior investigations into human reasoning have not sufficiently delineated its structure and components (Carey 2000; Hauser, Chomsky, and Fitch 2002; Guyer 2010). This study endeavors to fill this void by providing a detailed examination of LLMs within the ambit of the human reasoning framework, aiming to elucidate the nuanced interplay between AI capabilities and human reasoning processes.

In this research, we tackle the complexities of LLMs through the lens of Martin Heidegger's philosophical constructs of "ready-to-hand" and "present-at-hand" discussed in his work "Being and Time" (BT) (Heidegger, Macquarrie, and Robinson 1962) enriching our understanding of LLMs' roles within the domain of human reasoning. Heidegger's framework illuminates the dual nature of entities: "ready-to-hand" entities are appreciated for their practical utility, while "present-at-hand" entities are scrutinized for their intrinsic qualities from a scientific attitude. This duality forms the basis of what we term our *Faculty of Knowledge*, delineating the intricate structure by which we perceive and engage with our surroundings.

Our exploration extends to dissecting the "ready-to-hand" and "present-at-hand" dynamics, uncovering the statistical variations that contribute to the observed limitations in LLMs. We attribute these variations to the subjective and diverse modes of engagement with entities in their "ready-to-hand" state and the fluid nature of "present-at-hand" entities across various contexts. Our inquiry uncovers that "ready-to-hand" links can be qualified by their causal, dynamic, context-sensitive, and probabilistic nature, while "present-at-hand" links manifest as permanent, optional, or variable. Such distinctions enable us to map out the strengths and weaknesses inherent in LLMs.

Moreover, we tackle the pivotal question of how AIs based on LLMs compare with human reasoning. The initial step in this inquiry involves a careful examination of the human reasoning system, identifying all the sources of information that could be fed into the reasoning processes. For this examination, the traditional classifications of reasoning—whether by methodologies such as inductive, deductive, abductive, and analogical reasoning; by application domains like scientific reasoning; or by outcomes, namely explicative or ampliative reasoning—offer limited insights.

Psychological investigations, including the exploration of System 1 and System 2 thinking (Evans 2003; Kahneman 2011) and the differentiation between fluid and crystallized intelligence (Horn and Cattell 1967); (Ghisletta et al. 2012), shed some light on the subject. However, these models, grounded in empirical research, do not fully encapsulate the reasoning system, especially when it comes to non-verbal cognitive processes that are challenging to directly observe.

Given the complex and sometimes chaotic classifications of reasoning, it is impossible to initiate a structural analysis which will allow us to compare the nuanced aspects of human reasoning with the functionalities of LLM-based AIs, thus providing a clearer picture of AI's potential and limitations relative to human reasoning abilities. To overcome such difficulties, our inquiry into the structure of the human reasoning system is guided by Heidegger's concept of truth as "unconcealment,"(Wrathall 2010; Heidegger 2013) which posits that the discovery of truth involves both subjective perspectives and objective revelations to uncover the hidden aspects of entities. This approach allowed us to systematically delineate the inputs, outputs, and core structural elements of reasoning, drawing inspiration from Kant's detailed examination of the *Faculty of Understanding* in his work "Critique of Pure Reason" (CPR) *(Guyer 2010)*. Within this framework, we identify two essential inputs to the reasoning process: firstly, the guiding principles derived from our existing knowledge, embodying the subjective dimension; and secondly, the introduction of new observations about entities, facilitated by our guiding knowledge, representing the objective dimension. These inputs are deemed the sole avenues for the generation of new knowledge within our proposed reasoning model.

With these inputs clarified, our analysis categorizes the human reasoning system based on three criteria: the creation of new knowledge, the direct or projective/transformative use of pre-existing knowledge, and the incorporation of empirical observations. Each criteria requires support from a specific structural composition of distinct mind faculties. This structural-analytical division aims to chart the comprehensive architecture of the human reasoning system analytically, transcending the segmented understandings derived from psychological studies or empirical methodologies.

We have delineated four principal categories of reasoning: *Direct Explicative Reasoning*, *Projective Explicative Reasoning*, *Explorative Creative Reasoning*, and *Abstract Creative Reasoning*. *Direct Explicative Reasoning* entails the straightforward application of existing knowledge to the entity of interest without the intervention from intermediary cognitive faculties. *Projective Explicative Reasoning* demands the projection of the entity of interest into an abstract form, facilitating the application of theoretical knowledge to practical situations or the generalization of knowledge from the practical to the theoretical.

*Explorative Creative Reasoning* leverages existing knowledge for the possibility of creating new knowledge through two distinct strategies: *Transformative Creative Reasoning*, which involves mapping the entity of interest onto an unrelated entity to apply insights from this new perspective back to the original entity, and *Presentational Creative Reasoning*, where the entity is presented and observed in an novel coordinate system identified through existing knowledge. Note that it's important to distinguish between *Projective Explicative Reasoning* and *Transformative Creative Reasoning* in terms of their mapping methodologies and applications, as elaborated in the main text. Finally, *Abstract Creative Reasoning* focuses on the development of new concepts and theories grounded on empirical evidence. This type of reasoning is about pushing the frontiers of knowledge through abstraction, leading to innovative theoretical frameworks.

Our study of the human reasoning system aims to methodically uncover its core elements. We are not looking to restructure existing frameworks of reasoning but to highlight the key cognitive faculties foundational to different reasoning processes. For instance, we grouped syllogistic deductive reasoning, abstract problem-solving, and computer programming under *Projective Explicative Reasoning* because they all involve projecting a specific problem onto an abstract model and applying theoretical knowledge to solve real-world problems. The discovery of shared cognitive processes among different reasoning methods has allowed us to more accurately delineate the cognitive faculties involved and LLMs' role in these reasoning processes.

The discovered faculties include but are not limited to the *Faculty of Projection*, *Faculty of Judgment*, *Faculty of Analogy*, and *Faculty of Correlation*. This exploration is designed to pinpoint where LLMs might fit within a comprehensive reasoning paradigm. These faculties are conceptualized not as distinct physical entities within the brain but as cognitive units that may overlap in their functionalities. Our focus is not on the intricate internal makeup of these faculties but rather on understanding their inputs, outputs, and primary functions, which aids in identifying the hurdles in simulating these cognitive faculties in AI platforms.

In our discourse, we stick to traditional definitions of terms: We view 'concepts' as the foundational elements of our knowledge system, and within the realm of language, these concepts are expressed via words. The knowledge framework articulated through language includes both the descriptive and utilitarian characteristics of entities, as illustrated by the "ready-to-hand" and "present-at-hand" links. Moreover, observations are considered to be unprocessed information that has yet to be transformed into knowledge.

This framework equips us with the tools to analyze key structural elements of human reasoning, allowing for a comparison with their counterparts in LLM-based AI systems. Our analysis reveals that current LLM-based AI systems show proficiency in *Direct Explicative Reasoning* and some capability in *Projective Explicative Reasoning*. However, they have yet to master true *Rational Explicative Reasoning* fully and can only perform *Pseudo Rational Reasoning* without an AI counterpart of the *Faculty of Judgement*. At present, LLM-based AIs do not possess genuine creative reasoning capabilities, though initial steps towards *Transformative Creative Reasoning* appear promising.

Further, we evaluate the capabilities and potential risks of LLM-based AIs when they are combined with actively developed AI technologies such as graphical generative AI and embodied AI. Our analysis concludes that, at this stage, LLM-based AIs do not present an existential risk to humanity. Nonetheless, it is crucial to establish stringent regulations in the realm of creative reasoning research to ensure responsible advancements in this field.

This paper unfolds in the following manner: Initially, in Section 1, we delve into Martin Heidegger's philosophical constructs of 'Dasein's World' constructed from 'ready-to-hand' and 'present-at-hand' relational links, which together form our *Faculty of Knowledge*. We proceed to present the argument that LLMs serve as a digital counterpart to the human *Faculty of Verbal Knowledge*. Following this in Section 2, we explore Heidegger's notion of "unconcealment" as a foundational principle for reasoning, detailing the inputs and outputs of this reasoning framework. The subsequent Section 3 is dedicated to dissecting the structure and components of the reasoning system. We then conduct a thorough analysis of each reasoning category and sub-category, pinpointing their specific functions, mechanisms, and the essential mind faculties involved. Building on this foundation, we articulate the parallels between LLM capabilities and human reasoning faculties, thereby assessing the capabilities and limitations of LLMs in each category of reasoning. In Section 4, the discussion then extends to the risks and potentials of augmenting LLMs with generative AI and embodied AIs. The paper ends in a conclusion that synthesizes our findings and reflections.

**Section 1: Exploring Large Language Models in the Context of Heidegger's "Dasein's World"**

Martin Heidegger's philosophical framework, notably the idea of "Dasein's world," provides a meaningful lens to analyze and comprehend the capabilities and limitations of LLMs. In his seminal work, "Being and Time" (BT) (Heidegger, Macquarrie, and Robinson 1962), Heidegger introduces Dasein as a term that goes beyond traditional definitions of human existence found in fields like archaeology, biology, or psychology. At the core of Dasein is the concept of "care," which serves as the driving force behind all human activities. This framework can enrich our understanding of LLMs by examining how humans interact with entities in the world in a nuanced manner (BT, H194, pp.239).

> *In willing, an entity which is understood-that is, one which has been projected upon its possibility gets seized upon either as something with which one may concern oneself, or as something which is to be brought into its Being through solicitude. Hence, to any willing, there belongs something willed, which has already made itself definite in terms of a "for-the-sake-of-which". If willing is to be possible ontologically, the following items*

> *are constitutive for it: 1. The prior disclosedness of the "for-the-sake-of-which" in general (Being-ahead-of-itself); 2. The disclosedness of something with which one can concern oneself. (The world as the "wherein" of Being-already); 3. Dasein's projection of itself understandingly upon a potentiality-of-Being towards a possibility of the entity "willed". In the phenomenon of willing, the underlying totality of care shows through.*

Humans continually set future goals. After setting these goals, individuals interact with entities—whether objects or other people—that can aid in achieving these objectives. This phenomenon is termed "for-the-sake-of-which," a perspective that becomes apparent through a cognitive process that Heidegger calls "projection", which is also a characteristic of understanding.

Through the cognitive process of projection, entities acquire significance according to their functional utility, also known as their "readiness-to-hand." For instance, a hammer is principally valued for its capability to drive nails (BT, H69, pg. 98). Yet, when the hammer fails to meet its intended function, its status transitions from "ready-to-hand" to "present-at-hand." In this state, our attention may shift to its physical properties rather than its previous utility (BT, H73, pg. 103). Crucially, this transition is not static but fluid; a malfunctioning hammer might find renewed purpose as a paperweight, thereby reverting to a "ready-to-hand" status within a new situational context. These switching between "ready-to-hand" and "present-at-hand" states are intrinsically linked, subject to change based on specific circumstances and varying perspectives.

Natural entities, too, can be imbued with varying degrees of "readiness-to-hand," (BT, H71, pg. 100) depending on how they fit into different human activities or objectives. For instance, within the scope of home construction, trees take on a functional role as lumber—effectively becoming "tree-as-lumber" in our cognitive framework. Conversely, during a leisurely stroll in a park, the same tree might be conceived as a "tree-as-oxygen-provider" or a "tree-as-shade-provider." With each new context, the entity we identify as a "tree" forms different relationships, linking it to diverse actions and objectives. These functional links are typically pre-disclosed to our understanding and always exist within the framework of a larger "for-the-sake-of-which" objective.

The intricate weave of relationships, covering both utilitarian and descriptive qualities, creates a complex tapestry of interactions which collectively contribute to our understanding and interpretation of the "world."

> *The "wherein" of an act of understanding which assigns or refers itself for which one lets entities be encountered in the kind of Being that belongs to involvements; and this "wherein" is the phenomenon of the world.(BT, H86, pg. 119)*

Heidegger elaborates on "projection" as the cognitive act of choosing a specific functional role from a host of potential roles an object could fulfill. Language reflects these selected utilities. Taking the hammer as an example, a typical dictionary definition might describe it as "a hand tool featuring a heavy metal head affixed perpendicularly to a handle, used primarily for breaking things or driving nails." This linguistic description focuses on the core features of hammer's utility, while omitting ancillary features like the material of the handle. This illustrates how language, in naming the hammer, adopts a selective and subjective utility-focused perspective. This practical attitude isn't limited to artificial objects but also applies to natural ones. For instance, the reason we refer to certain black, juicy berries as blackberries, but not others, stems from a utilitarian perspective as well: their nutritional value, taste, and ease of growth. Such naming reflects a human-centric valuation, emphasizing traits that benefit us (Wrathall 2010)

Although the attitude we adopt in present-at-hand inspections is not utilitarian, it is nonetheless still subjective and selective as soon as we articulate the presence-at-hand in concepts of language. Because concepts are always composed of a limited set of features, when we articulate an entity in its presence-at-hand, we cannot avoid describing the entity only by a subset of its total features encapsulated in the concepts of description.

There's a tacit aspect to both "readiness-to-hand" and "presence-at-hand" that exists outside the realm of verbal description. For example, the sensory experience of holding a hammer and the bodily movement involved in using it for work often remain unspoken and absent in linguistic definitions. As to present-at-hand descriptions, while an entity can be described by a collection of predicative terms, some aspects are always missing by the predicates. As a result, words are not wholly rooted in their utilitarian or descriptive features; they also serve as a referential signpost pointing to the real entities. This perspective has long fueled arguments that non-embodied AIs such as LLMs are incapable of understanding language in its fullest, truest sense (Dreyfus 1992).

However, as the analysis of the concepts of "ready-to-hand" and "present-at-hand" have revealed, the scope of language extends beyond mere literal references to physical entities. The essence of language is also shaped by a complex web of relational links that are imposed on the physical entities by our utilitarian attitude and the unavoidable selectiveness in present-at-hand descriptions. Such a web entails the knowledge of how things can be utilized for achieving various goals and it is an integral part of human decision making.

Understanding LLMs through Heidegger's lens of "Dasein's world" offers valuable insights into how these models comprehend and generate language. These insights can aid in deciphering both the capabilities and limitations of LLMs, thus providing a more nuanced understanding of human interaction with AIs. By employing Heidegger's intricate philosophical frameworks, we can delve deeper into the essence, functionalities, and limitations of LLMs, enriching the discourse surrounding technology and human reasoning.

## 1.1 Exploring Entity Relationships Through Heidegger's Concepts of Ready-to-Hand and Present-at-Hand

### 1.1.1 Characteristics of Ready-to-hand Relational Links

Recognizing the essence of an object's "readiness-to-hand" is vital, as it is deeply influenced by the characteristics that render the object valuable in achieving our specific aims and desires upon interaction. This concept transcends a mere depiction of an object, spotlighting instead its practical importance within a vast web of relationships and activities. Being fundamentally relational, "readiness-to-hand" is shaped by the purpose it serves, a concept Heidegger refers to as "for-the-sake-of-which." For example, when employing a hammer to insert a nail, the hammer doesn't just interact with the nail but becomes a part of a larger sequence of actions directed towards a broader aim, such as building furniture. In this framework, each "ready-to-hand" link signifies a connection from cause to effect through specific actions, illustrating how a sequence of such connections can accomplish complex tasks.

The nature of an object's "ready-to-hand" quality is flexible, adjusting to the evolving demands and circumstances. The function of a hammer, for instance, extends beyond merely driving nails; it can also be employed for breaking materials apart among other uses. This illustrates the flexible essence of "readiness-to-hand," which accommodates the various functional capacities an object can assume, each tailored to distinct purposes. In this light, objects are not simply encountered in their inherent state but are rather identified and molded into specialized tools within the contexts of our interactions with them.

In scenarios where an object can serve various purposes, the probability of a particular ready-to-hand link being activated hinges on the specific intention behind the action. Take the example of a hammer: when the aim is to disassemble furniture, the function of prying becomes crucially important, in contrast to when constructing furniture, where prying may not be necessary. This variability in usage can be expressed through statistical terms. Generally, a hammer might be frequently used for hammering, with prying being a less common action, making hammering more probable than prying. Yet, in contexts such as furniture disassembly, the likelihood of using the hammer for prying increases markedly. This shift in probabilities can be effectively represented through conditional probability, as the increased likelihood of prying when the task is furniture disassembly. Broadly speaking, the more information available about the task at hand, the higher the probability of selecting the appropriate "ready-to-hand" function for the job.

To summarize, the concept of "readiness-to-hand" encompasses several critical characteristics: it is goal-oriented, reflecting its basis in utility; dynamic, as it necessitates action; causal, due to its capacity to bring about specific outcomes; flexible, since one entity can be used in multiple ways; context-sensitive, recognizing that an entity can assume different utilitarian roles pertinent to different scenarios; and statistical, acknowledging that the likelihood of employing different utilities fluctuates with the context.

### 1.1.2 Characteristics of Present-at-hand Relational Links

Present-at-hand serves as a theoretical framework for understanding entities through predicative descriptions, often articulated in terms of their inherent characteristics or dynamic behaviors. This cognitive attitude assumes a secondary role when "readiness-to-hand," or functional utility, is the primary concern. However, it moves into the foreground when utilitarian considerations recede. Presence-at-hand can manifest either as explicit presence-at-hand defined through predictive descriptions or implicitly grasped features that have not been captured in language.

Both explicit and implicit characteristics of an entity, as observed from a "present-at-hand" perspective, can lead to discoveries of new utilities for that entity. Consider a tree that is deemed too gnarled to be of use as lumber (Chuang 2000). The attribute—gnarliness—represents the tree's "present-at-hand" qualities that are useless for its utility as lumber. Yet, these same perceived drawbacks can be reevaluated as advantages in the context of seeking a shade tree. A tree's twisty form, which may render it undesirable for lumber, ironically makes it an ideal candidate for providing shade, as it is less likely to be cut down. In this way, these "present-at-hand" features of the tree are reassessed, leading to the recognition of its value in a new utility.

Present-at-hand characteristics can be categorized based on their stability. Certain traits are inherent to an entity, acting as its signature attributes, whether it's a natural or man-made object. For instance, a hammer is inherently known for its striking head, just as a mountain is recognized by its height. These attributes establish permanent descriptive connections to the entity. On the other hand, some characteristics are less consistently linked; for example, a hammer may not have a rip claw and a mountain may not have cliffs. These aspects create optional connections with the entity. Additionally, certain traits are always probabilistically tied to an entity, reflecting conditions that can change over time. Water, for instance, may transition between steam and ice based on the temperature. Thus, present-at-hand characteristics can be classified as permanent, optional, or variable (statistical) based on their relationship with the entity.

Like the way practical "ready-to-hand" links require context for clarity, "present-at-hand" connections also benefit from contextual detail: The inclusion of more contextual information refines our understanding of an entity: we gain a more detailed and accurate insight into the characteristics or states of entities, akin to how additional context enhances our grasp of an entity's practical use in "ready-to-hand" scenarios. Articulating these relationships within a well-defined context through language leads to a deeper comprehension of the meanings of words, facilitating clearer communication and interpretation.

## 1.2 Ready-to-hand and Present-at-hand Links Entails the Essence of Language

Heidegger suggests the view that language is a unifying force, bringing together, making accessible, and preserving the relationships among things and the entities themselves, thereby governing their interaction as stated in "Aristotle's Metaphysics Theta*"* (AMT) ((Heidegger 1995)

> *"...Language is, as saying, that forms the world's ways, the relation of all relations. It relates, maintains, proffers, holds, and keeps them" (AMT: 107)*
> *"...Such a gathering, which now gathers up, makes accessible and holds ready the connections of what is connected, and with this the connection itself and thus individual entities, and so at the same time lets them be governed, this is the structure that we call language". (AMT: 121)*

This conceptualization of language as a structure that binds and reveals the essence of connections and entities represents a profound shift from traditional linguistic theories that view language as "a stock of individual terms and rules for linguistic construction" as summarized by Heidegger in "Elucidations of Holderlin's Poetry" (EHP: 39) (Heidegger and Hoeller 2000).

Relational links offer a parsimonious approach to linguistic depiction, adaptable across various syntactic structures, thus accommodating the richness of linguistic variation. For instance, the relationships among a hammer, the act of hammering, and a nail can be depicted using different grammatical voices and clause structures, such as active or passive voice, and main or subordinate clauses. Past research, such as studies by Psarologou et al.(Psarologou, Esposito, and Bourbakis 2015; Mills 2013) and Mills (Psarologou, Esposito, and Bourbakis 2015; Mills 2013), maps text onto Statistical Petri Nets (SNPs), a specialized form of relational graph which underscores the idea that the web of connections between words—captured via relational links—serves as a more fundamental layer of meaning than the intricate syntactic structures often found in sentences.

## 1.3 Ready-to-hand and Present-at-hand Links Constitute *the Faculty of Knowledge*

The complex web of "present-at-hand" and "ready-to-hand" links underpins our comprehension of entities, delineating their properties, functionalities, and how they interact, thereby forming the essence of our knowledge system. This knowledge is divided into two primary categories: *Verbal Knowledge* and *Experiential Knowledge*. Within this discussion, verbal knowledge refers to information that can be expressed or communicated via language, including both oral and written forms. Experiential knowledge, on the other hand, is characterized by its non-linguistic outputs, manifested through actions aimed at achieving specific goals.

In the *Faculty of Verbal Knowledge,* the structure consists of nodes that represent concepts, interconnected by links that define the relationships between these concepts. This network is the foundation for how knowledge is organized and accessed: when prompted, specific connections within this network are activated based on the content of the prompt, altering the likelihood of their activation. This mechanism implies that the network's configuration, or the state of knowledge, is determined by the conditional probability of activating these links given the prompt. The activation process continues until either all highly probable links are activated, or the capacity of working memory is reached, at which point the retrieval of information concludes.

The *Faculty of Experiential Knowledge* holds a comprehensive repository on which actions to take in various environments, aligned with human intentions. This faculty's ability to be activated by non-verbal cues underscores its autonomy from the *Faculty of Verbal Knowledge*. Nevertheless, the fact that it can operate in conjunction with the *Faculty of Verbal Knowledge* indicates a reciprocal link between them. Despite its non-verbal nature, when a utilitarian approach is employed, specific features of entities are emphasized, while others are diminished, which paves the way for the symbolic representation of such relational links through means such as gestures or contextual clues, establishing a "norm" of interaction with objects. These norms are made up of "control parameters" that direct our physical interactions with tools and the manipulation of our surroundings. It's important to recognize that the norms inherent in our interaction with the world are predominantly conveyed through action rather than speech. As Wrathall (Wrathall 2010) (pg. 54-55) points out, "The normativity inherent in our engagement with a world is usually transmitted practically rather than linguistically." "It is thus on the basis of our pragmatic discovery of things that language is possible, for it is the structure of equipment and involvement built into our comportment that delineates the features of things that are salient to us- the very features that form the content of our beliefs and utterances". In other words, the practical grasp of entities as tools forms the basis for their linguistic presentation.

Beyond the *Faculty of Knowledge* lies the closely related *Faculty of Observation (or Reproductive Imagination in Kantian terms)*, which is triggered by environmental cues, highlighting its autonomy from the broader *Faculty of Knowledge*. This cognitive faculty, stored within memory, autonomously records observations without active human engagement, capturing entities in their entirety without the selective focus required for conceptualization that establishes relational connections among entities. Consequently, observations resist symbolic representation due to their comprehensive and non-selective nature, being intrinsically non-relational. Nevertheless, the *Faculty of Observation* plays a vital role in deepening our understanding, suggesting a reciprocal connection with the *Faculty of Knowledge*. This indicates that although observations themselves do not categorize entities through utilitarian or descriptive relationships, their integration with knowledge—which does involve such connections—fills this void. Ultimately, knowledge and observations represent the explicit and implicit aspects of the present-at-hand and ready-to-hand links respectively.

It's important to recognize that the knowledge embodied in the concepts of "ready-to-hand" and "present-at-hand" is often acquired through learning or adherence to societal norms, reflecting how Heidegger's concept of "Das Man" or the "they" typically engages with the world's entities. According to Heidegger (BT, H129, pp167)

> *When entities are encountered, Dasein's world frees them for a totality of involvement with which the "they" is familiar, and within the limits which have been established with the "they's" averageness…*
> *Proximally Dasein is "they", and for the most part it remains so. If Dasein discovers the world in its own way [eigens] and brings it close, if it discloses to itself its own authentic Being, then this discovery of the 'world' and this disclosure of Dasein are always accomplished as a clearing- away of concealments and obscurities, as a breaking up of the disguises with which Dasein bars its own way.*

In this analysis, Heidegger argues that the customary ways Dasein interacts with the world—through "ready-to-hand" and "present-at-hand" links—are reflective of the collective norm, "Das Man." He sets this routine interaction against the notion of authenticity, achievable only by stripping away the layers of concealment and obscurities. Thus, it follows that these layers of concealment and obscurities are products of "Das Man's" typical modes of engaging with the world, modes that are structured around "ready-to-hand" and "present-at-hand" links. This perspective also opens up the possibility for the existence of personal and distinct relational links that stem from an individual's authentic self, even though most links are learned from societal conventions. A more proximal analysis reveals that all such relational links inherently focus on certain features of entities over others, thereby obscuring their real nature. A genuine understanding of entities in their real form is only possible when these constructed relational links, whether adopted from society or crafted by oneself, are removed. This allows an individual's authentic self to directly perceive entities in their real form. Consequently, we can categorically state that all relational links found within the *Faculty of Knowledge* are essentially inauthentic, as they fail to represent entities as themselves but only represent them by sets of selective features.

## 1.4 LLMs are Computational Analogues of *Faculty of Verbal Knowledge*

LLMs are engineered to statistically understand the relations between words. These models convert input sequences into high-dimensional, vector-based representations that capture complex relationships between words (Brown et al. 2020; Vaswani et al. 2017). The training process initiates by embedding a set of prompting words as tokens, each stamped with positional encodings. Utilizing the attention mechanism, the model estimates how each word in the prompt contributes to the prediction of the subsequent word, translating these contributions into attention scores. The information then passes through a fully connected layer, transforming the input into vectors that encapsulate both the characteristics of the prompting words and their surrounding context. To refine this, LLMs often employ multiple layers of transformers to accumulate increasingly nuanced relational and contextual data, thereby enhancing the accuracy of the predicted output word.

Over time and with extensive training data, LLMs can identify and catalog a myriad of connections between existing words, forming a statistical tableau that represents the joint probability of all word tokens. Given a specific set of prompting words and a conversational history, the model can compute conditional probabilities for potential responses, which are represented as attention scores based on the trained weights of neural connections.

Although constructed through statistical means, the resulting structure from LLM training transcends mere data, serving as a complex, vector-based portrayal of linguistic interconnections. This portrayal, despite its mathematical nature, encapsulates ready-to-hand and present-at-hand relationships among words, phrases, and broader linguistic contexts. For example, the vector representation of the word "king" is multi-dimensional. Some dimensions may highlight ready-to-hand relationships, such as the synaptic construct between "king," "rules," and "kingdom". Simultaneously, other dimensions may emphasize present-at-hand relations, linking "king" to royalty-associated terms like "queen," "prince," or "duke," or even contextual relationships like the word "king" often appearing in historical or fantasy narratives (Heimann and Hübener 2023).

*The Faculty of Verbal Knowledge* comprises a complex network of relational links, and LLMs like ChatGPT are engineered to map these connections through weighted relationships between word tokens. Therefore, LLM-based AI systems can be considered linguistic approximations of the *Faculty of Verbal Knowledge* which encapsulate relational links between words, offering nuanced approximations of Heidegger's concepts of readiness-to-hand and presence-at-hand.

By comparing LLMs as computer analogues of the *Faculty of Verbal knowledge*, we can see that LLMs are deficient in two ways. First, LLMs lack an equivalent mechanism for accessing the *Faculty of Experiential Knowledge*. Second, their disconnection from the *Faculty of Observation* narrows their understanding to the confines of predefined linguistic networks. This deficiency in experiential insight curtails their ability to apply knowledge in practical, real-world scenarios. Additionally, the absence of a connection to observational capabilities restricts their functionalities that depend on direct sensory engagement or observation.

LLMs, like Heidegger's concept of "Das Man " or the "they," which are reliant on verbal relational links, are categorically inauthentic. However, equating LLMs directly with "Das Man" warrants further scrutiny. "Das Man "is characterized by features such as distantiality, averageness, leveling down, and publicness" (BT H129, p.166). Considering LLMs do not possess emotions, attributing distantiality to them would be misleading, as they do not differentiate themselves from others. Although LLMs develop their relational links through training, stating that these links represent "publicness" and "averageness" might oversimplify their complexity. LLMs aggregate relational links from a broad spectrum of inputs, possibly including distinctive ones that do not conform to the most common patterns. Thus, it's arguable that LLMs mirror "Das Man" to a significant extent, yet they don't entirely eliminate unique nuances in their learning process. As a result, it's still possible to guide LLMs to emulate specific persona in their text output.

Armed with this enhanced understanding, we are better positioned to address perplexing questions concerning the capabilities and limitations of LLM-based AI. Our subsequent discussions will focus on key questions outlined in the introduction, shedding light on this intriguing subject matter.

**Section 2: A Structural Sketch of the Reasoning System for Elucidating Capabilities and Limitations of LLMs**

LLMs have demonstrated the ability to embody the structures and functionalities of "ready-to-hand" and "present-at-hand" link networks, showcasing their potential to emulate human reasoning. Despite these insights, a detailed exploration into their operational strengths and limitations is yet to be conducted. Notably, Heidegger's contributions did not address how these networks correlate with broader aspects of cognition and reasoning. Additionally, there is a clear gap in establishing a methodical framework for examining the role of LLMs within the greater scheme of human reasoning. This framework should specify the cognitive faculties involved, their interactions, and a detailed classification of reasoning types from a structural viewpoint. This viewpoint emphasizes the use of input data to obtain insights about a subject, because varied applications of inputs require different foundational structures. This approach diverges from conventional reasoning classifications, which are organized by application area (e.g., scientific reasoning), methodology (e.g., deductive, inductive, abductive), goal orientation (e.g., speculative, or practical), and scope (e.g., explicative or ampliative), focusing on whether the reasoning clarifies existing premises or extends beyond them.

In modern psychology, distinctions of reasoning are mainly drawn from the functional perspective rather than the structural perspective. For example, human reasoning is frequently divided into two types: System 1 and System 2 thinking, as identified by (Evans 2003; Kahneman 2011). System 1 operates on an intuitive and reflexive basis, drawing on past experiences automatically without deliberate thought. In contrast, System 2 involves deliberate analytical thinking, employing logical reasoning and detailed evaluation of previous knowledge and experiences. In addition to this distinction, the dichotomy of fluid and crystallized intelligence, introduced by Horn and Cattell (Horn and Cattell 1967) and explored further by Ghisletta et al. (Ghisletta et al. 2012), emerges from linear factor analysis. Fluid intelligence encompasses the capacity for abstract thinking and problem-solving independent of learned knowledge, whereas crystallized intelligence builds on the foundation of accumulated knowledge and experiences, typically strengthening with age as described by Horn (Horn 1968). However, recent studies, such as those by Huang and Zhang (Y. Huang and Zhang 2022), have pointed out the constraints of traditional linear factor analysis in fully capturing the nuances of these intelligence types. These psychological categorizations arising from empirical observations are often fragmented.

In the complex field of reasoning, there's a pressing need to systematically understand the reasoning system's structure and function. The first step towards this inquiry involves outlining the system's inputs and outputs and identifying its key components, a task reminiscent of Kant's exploration of the *Faculty of Understanding* (CPR, B93), where he delineated sensory intuition and concepts as inputs, with cognition of understanding as the output. However, we still lack a point of departure for elucidating the inputs and outputs of the reasoning system.

Kant assigns to reason the task of seeking unity and systematization of knowledge obtained through understanding, a process that involves both analytic and synthetic judgments. While analytic judgments clarify concepts by elucidation, it's through synthetic judgments, particularly synthetic a priori judgments, that Kant sees the potential for generating new knowledge. These judgments are crucial in fields like mathematics and physics, where they enable universal and necessary knowledge independent of empirical experience. However, Kant's emphasis on synthetic a priori judgments as the model for generating new knowledge has been critiqued for its perceived limited applicability beyond these domains (CPR, Introduction). Also, important reasoning types such as analogical reasoning were missing from his discussions. Nonetheless, Kant's work underscores the necessity of a clear definition of reasoning as a foundational step toward fully illuminating the reasoning system, warranting a dedicated discussion on this topic.

## 2.1 The Essence of Truth as the Ground for Defining Human Reasoning.

The preceding discussion has revealed that traditional definitions of reasoning fall short for our investigation into the structural aspects of the human reasoning process. This necessitates wiping the slate clean and beginning anew. In general terms, reasoning is understood as a cognitive process aimed at uncovering what is not immediately apparent. If something is clear and obvious, reasoning is not required. While reasoning is often considered within the realm of epistemology, it is crucial to recognize that new knowledge can emerge from direct observation, bypassing the reasoning process. This suggests that, despite some overlap, reasoning and epistemology are distinct concepts. Additionally, reasoning is typically separated from the creation of art and artificial tools in our collective understanding. Kant positions reason as a subsequent stage to understanding, proposing a linear progression from sensory perception to understanding, and finally to reasoning. This perspective overlooks the influence of reasoning on shaping our observations. However, it is evident in scientific inquiry and other activities that we constantly engage in reasoning to devise new observational methods. These initial considerations demonstrate the challenge in articulating a definition of reasoning that accurately captures its essence.

Amid these confusions, the notion of truth as "unconcealment," proposed by Heidegger, emerges as a fitting foundation for our discussion. Heidegger suggests:

> *To say that an assertion "is true" signifies that it uncovers the entity as it is in itself. Such an assertion asserts, points out, 'lets' the entity 'be seen' in its uncoveredness. The Being-true (truth) of the assertion must be understood as Being-uncovering* (BT: H219 pg.261)

This concept of truth introduces a subjective and investigative aspect: the quest for truth is an active process of shedding light on aspects of reality that were previously obscured. By embracing the view that reasoning seeks to uncover hidden truths, the process of reasoning begins with an attempt to reveal the concealed, and through the insights gained under this new light, we engage in reasoning to clarify our understanding and, possibly, to generate novel knowledge.

Concepts are described metaphorically in "unconcealment": what is hidden refers to aspects of entities that are not immediately visible, and the act of bringing them to light is a subjective process aimed at rendering these entities observable. Unlike the formation process of sensory intuition, where sensory organs passively receive light, the metaphorical 'light' in this context is actively projected onto entities by the subject, and it can be nothing else but the sum of human knowledge that can guide us toward new discoveries. Therefore, in this formulation, reasoning is driven by two main elements: the observation of the entity of interest and the total collection of existing knowledge.

The outcome of the reasoning process is the elucidation of knowledge concerning the entity in question. This revelation does not necessarily involve the discovery of new knowledge about the entity; even the recollection or clarification of known knowledge is considered an act of "unconcealment."

By framing "reasoning" as a process of "unconcealment," this definition aligns with the criteria we have established for reasoning: it reveals what is concealed, thereby facilitating the discovery of new knowledge without necessitating that the revelation itself constitutes new knowledge. Furthermore, this approach addresses the omission in Kant's theory regarding the transition from reasoning to new observations by highlighting that the act of illumination is inherently propelled by reasoning.

**2.2 An Initial Sketch of the Structure of the Reasoning System**

In our prior discussions, we've outlined the inputs and outputs of the human reasoning system. This groundwork paves the way for us to explore the components and framework of such a system. While a comprehensive examination of this intricate subject is beyond our current scope, we can construct a preliminary model of its architecture. This initial model will enable us to assess the potential capabilities and limitations of LLMs.

Given that we want to elucidate the structure of the reasoning system, we can first categorize reasoning into two main types: *Non-creative* and *Creative Reasoning*. *Non-creative Reasoning*, also known as *Explicative Reasoning*, operates within the confines of the existing knowledge base, elucidating what is already known without contributing new knowledge. This type of reasoning does not necessitate structural elements for expanding the knowledge base as in *Creative Reasoning*.

*Explicative Reasoning* splits into two distinct types: *Direct* and *Projective Explicative Reasoning*, differentiated by their way of application of knowledge. *Direct Explicative Reasoning* straightforwardly retrieves knowledge concerning the entity of interest to disclose its concealed aspects. In contrast, *Projective Explicative Reasoning* involves projecting the source entity to an abstract model, enabling the application of knowledge either from the abstract model to the source entity or vice versa. These approaches have unique structural prerequisites: *Direct Explicative Reasoning* operates without reliance on additional cognitive processes, whereas *Projective Explicative Reasoning* depends on the involvement of other mind faculties.

For creative reasoning, we categorize it based on whether the process relies on the subjective manipulation of known knowledge, or the objective analysis of empirical data related to the entity of interest. This distinction leads to two types: *Explorative Creative Reasoning* and *Abstract Creative Reasoning*. *Explorative Creative Reasoning* utilizes existing knowledge either to develop new coordinate systems for presenting new observations of the entity of interest, termed *Presentational Creative Reasoning*, or to transform knowledge from one entity to another, known as *Transformative Creative Reasoning*. This form of reasoning seeks novel perspectives on the entity of interest when there is a lack of direct observation of it. *Abstract Creative Reasoning*, on the other hand, starts with empirical observations with the goal of distilling these into new abstract concepts or theories. This approach is generative, leading to the development of new theoretical knowledge that other reasoning processes rely on.

Building on the idea that truth fundamentally involves revealing what is hidden, we've pinpointed two critical inputs for the reasoning process: observations of the subject matter and the collective body of total knowledge. From there, we categorized the reasoning process into *Creative* and *Non-creative* segments, based on their structural differences. *Non-creative*, or *Explicative Reasoning*, splits further into *Direct* and *Projective Explicative Reasoning*. Similarly, Creative Reasoning divides into *Explorative* and *Abstract Creative Reasoning*, distinctions based on the application of existing knowledge.

Having outlined the framework of the human reasoning system, our next step involves a detailed exploration of each reasoning category and its subdivisions, assessing their inherent strengths and weaknesses. We'll then evaluate how LLMs align with each of these reasoning types, aiming to define the scope and constraints of reasoning as executed by LLMs.

**Section 3: Analysis of Various Categories and Subcategories of Human Reasoning and LLM's Counterpart in Them**

**3.1 Direct Explicative Reasoning**

Direct reasoning is defined as the act of retrieving relevant knowledge about an entity of interest directly in response to a prompt, which could be in various forms such as text, images, videos, or a blend of these media.

This straightforward approach to reasoning bypasses the need for any intermediary cognitive faculties, rendering it a reflexive form of reasoning that operates without the need for self-awareness. Yet, the insights gained from this direct engagement with our knowledge faculty are subsequently fed into our self-consciousness, enabling us to consciously comprehend the outcomes of our inquiry.

By examining "ready-to-hand" and "present-at-hand" links following a prompt, we can piece together an entity's previous state and predict its forthcoming scenarios. "Present-at-hand" links represent the current state, from which we can extrapolate past situations and forecast future developments, thanks to "ready-to-hand" links that connect causes to effects. Moreover, these relational links are sensitive to context, with the activation of specific links shedding light on the prevailing context. By analyzing the context alongside the activated "ready-to-hand" links, we can infer the overall intention of a prompt. Therefore, *Direct Explicative Reasoning* enables us to access hidden information, such as an individual's intention, by utilizing applicable existing knowledge.

Providing additional context within the prompt can further refine which relational links are activated. For example, by posing the question, "When taking things apart, what can we do with a hammer and nails?" The specific context of "taking things apart" selectively triggers "ready-to-hand" links that form a logical sequence from start to finish. In such scenarios focused on disassembly, links related to the use of the hammer for prying become more relevant. As more detailed context is included, these links become increasingly precise, enhancing the specificity and accuracy of the reasoning process. In addition, the activation of specific relational links within the web of relational links is contingent on the prompt containing sufficient detail to elevate these links above the threshold required for their significance in conditional probabilities. When a prompt offers only scant information, it results in the activation of a restricted set of relational links. This phenomenon leads to what psychologists have identified as the "What You See Is All There Is" (WYSIATI) fallacy, a common bias in System 1 thinking, where decisions are made based on the limited information immediately available.(Kahneman 2011).

Due to the lack of judgment in *Direct Explicative Reasoning*, the knowledge retrieved may include factual inaccuracies and inconsistencies. Additionally, the activation of links depends on statistical likelihood, meaning that a non-specific prompt might lead to a preference for more frequently occurring connections, introducing a bias towards conventional associations. Conversely, an improper prompt could activate unrelated links, resulting in an irrelevant response. Also, if the system has been activated before, the outcome of a current inquiry could be influenced not just by the present prompt but also by previously activated nodes and links, leading to varied responses to the same prompt over time. Lastly, given the limitations on how much knowledge can be accessed at once, a query might not reveal all pertinent information in one go, potentially omitting relevant details.

The characteristics of *Direct Explicative Reasoning* align closely with what psychology terms *System 1 thinking*. This mode of cognition is marked by swift mental operations that draw upon immediate concepts, familiar associations, and emotional responses, as observed and discussed in a variety of studies (Tulving, Schacter, and Stark 1982; Caruso, Shapira, and Landy 2017; Berger, Meredith, and Wheeler 2008; Vohs, Mead, and Goode 2006; Bolte, Goschke, and Kuhl

2003). System 1 thinking often glosses over nuanced information and exhibits a bias towards familiar concepts that are cognitively easier to process (Oppenheimer 2006; Kahneman 2011). It is prone to errors and tends to adhere to established norms (Oppenheimer 2006; Kahneman 2011; Kahneman and Miller 1986). This mode of thinking is largely automatic and unconscious, frequently identifying patterns and causal relationships to construct coherent narratives without deep reflection ((Michotte 2017; Hsu 2008); (Mednick 1962; Hsu 2008)). Given these parallels, *System 1 thinking* can be considered the same as *Direct Explicative Reasoning*.

### 3.1.1 Capabilities and Limitations of Direct Explicative Reasoning in LLMs

The architecture of our knowledge system, structured as a web of interconnected links, inherently shapes its functional reach and constraints when employed directly. LLMs act as a reflection of this system's verbal aspects, embodying many features of *Direct Explicative Reasoning*. For example, LLMs can understand a user's intent based on a prompt through the same mechanism as we have explained in section 3.1. GPT4 (Jan. 20, 2024) illustrated this capability in this dialogue:

> *Prompt:*
> *If I am looking for a nail box and a hammer, what am I intending to do?*
>
> *GPT4:*
> *If you are looking for a nail box and a hammer, it's likely that you intend to engage in some form of construction or repair work, such as hanging pictures, assembling furniture, or fixing something in your home or workplace that requires nails to be driven into a surface.*

However, a shortcoming in LLMs is their inability to tap into experiential knowledge, which includes the physical experiences associated with entities, thus precluding them from performing what is known as embodied reasoning. This limitation underscores the need for an integrated system that combines LLMs with a general embodied AI framework capable of processing such experiential knowledge. For instance, while an LLM might identify the connection between using a hammer and nails in response to a construction-related prompt, it lacks the understanding of the physical act of hammering.

In contrast, human reasoning seamlessly merges verbal guidance with non-verbal, experiential insights, such as the tactile knowledge of swinging a hammer, leading to a richer, more nuanced understanding. This integration of verbal knowledge with embodied experience as well as observations in human reasoning can initiate further explorations and insights, enhancing the depth of understanding in ways LLM-based reasoning currently cannot achieve due to its lack of access to a system equivalent to the *Faculty of Experiential Knowledge and Faculty of Observation.*

In the verbal part, shortcomings of *Direct Explicative reasoning* can find direct analogues in the behavior of LLMs. For instance, LLMs generate text, answers, or references when uncertain; commit obvious factual errors; and sometimes struggle with basic logical inferences or simple mathematical problems (Floridi and Chiriatti 2020);(Cobbe et al. 2021); (Perez and Ribeiro 2022); (Arkoudas 2023b); (Borji 2023). LLMs also exhibit the WYSIATI fallacy.

Consider an example from Chapter 10 of "Thinking, Fast and Slow" by Kahneman (Kahneman 2011). When people hear that certain rural counties have the highest rates of pancreatic cancer, they may hastily conclude that higher alcohol consumption in these areas is the cause. However, this overlooks the fact that some rural counties also have the lowest rates of pancreatic cancer, a phenomenon that could be attributed to statistical variations in small populations. *System 1 thinking (Direct Explicative Reasoning)* often falls prey to the WYSIATI fallacy. Similarly, LLMs like GPT-3.5 focus on the activated prompt words and their immediate associations, neglecting other potentially relevant information. Intriguingly, when posed with the same question about pancreatic cancer rates in rural areas, GPT-3.5 also suggested alcohol consumption as one of many plausible explanations (GPT-3.5, Aug. 10, 2023).

Despite these similarities, LLMs do have an advantage over human *Direct Explicative Reasoning* in terms of the breadth of their relational links across various domains. This enables them to generate a wider array of plausible cause-and-effect theories for a given phenomenon. For example, in the aforementioned conversation, GPT-3.5 offered 10 potential reasons for the observed cancer rates. Thus, while LLMs may be more comprehensive in generating hypotheses for further investigation, they share the same limitations of shallow reasoning inherent to *Direct Explicative Thinking.*

LLMs also stand out for their adeptness at language translation, setting them apart from typical human translation techniques, which can be categorized into two main approaches. The first approach involves forming sentences within the structure of the native language and then replacing words with their counterparts in the target language, often necessitating a rearrangement of word order as a subsequent adjustment. This method may result in translations that lack the smoothness and authenticity characteristic of the target language. The second approach is to translate directly using the relational structure of the target language, which tends to yield translations that sound more natural.

Mastering a foreign language to the extent necessary for fluent translation usually requires significant study, leading many people to rely on the first method for translation tasks. In contrast, LLMs are engineered to perform translations directly in the target language, drawing on their comprehensive training databases. Therefore, when they are adequately trained, LLMs have the potential to produce translations that better capture the essence and nuances of the target language although they may still struggle with context, cultural nuances, and idiomatic expressions to some extent, areas where a skilled human translator with deep cultural and linguistic knowledge might excel. Additionally, LLMs are proficient in converting everyday language into computer code or abstract concepts when correctly trained, making them an exceptional resource for navigating among various modes of expression.

### 3.2 Projective Explicative Reasoning

*Projective Explicative Reasoning* consists of mapping the entity of interest onto an abstract model and then applying theoretical knowledge to this entity or extending knowledge from the entity to its abstract model. The primary goal of this reasoning approach is to clarify existing knowledge, focusing on elucidation rather than the discovery of new insights. As such, it falls under the category of explicative reasoning.

*Deductive Syllogistic Reasoning* is a type of *Projective Explicative Reasoning* which starts with an entity of inquiry represented by a set of conceptual rules E={e1, e2..., en}. The task then is to identify a broader category G={g1, g2..., gm} that encompasses E, allowing conclusions about E to be deduced based on knowledge of G. However, upon closer examination, the process often described as generalization is more accurately termed abstraction. This is because for deduction to hold—where G is universally true if E is true, thereby including E within G—G must have a smaller set of conceptual rules than E, which aligns with the concept of abstraction. In common usage, generalization doesn't imply the reduction of features, making "abstraction" a more fitting term in the context of defining deductive reasoning.

In this formulation of deductive reasoning, the role of the *Faculty of Projection* is to identify an abstract form (or general category) G with characteristics {g1, g2…,gm} that are relevant to the question of inquiry. Subsequently, the *Faculty of Judgement* evaluates whether G indeed is a subset of E, allowing the application of G's knowledge to E.

Consider the question of whether mercury conducts electricity, approached through syllogistic deduction as an example. The *Faculty of Projection* formulates the inquiry in terms of mercury and electrical conductivity, projecting these onto the abstract category of metals. The *Faculty of Judgement* then assesses whether the category of metals is applicable to mercury. Affirming this connection means the knowledge that all metals conduct electricity can be applied to infer that mercury does indeed conduct electricity. If, alternatively, mercury were initially projected onto semiconductors, the *Faculty of Judgement* would dismiss this projection because mercury does not fit the abstraction of a semiconductor. This example illustrates how deductive reasoning involves a coordinated, iterative interplay among the *Faculty of Projection, Judgement, and Knowledge*.

*Inductive Syllogistic Reasoning* parallels *Deductive Syllogistic Reasoning* in its initial step of mapping an entity of interest, E, onto an abstract counterpart, G. The distinction lies in the directionality of applying knowledge: in this case, insights from E are generalized to G, and the reliability of this generalization may be quantified by a probabilistic measure. This method is also included within the larger framework of *Projective Explicative Reasoning.*

Abstract problem-solving is another type of *Projective Explicative Reasoning*. When faced with a problem concerning a particular entity, the *Faculty of Projection* abstracts the problem into a form that has known solutions. The initial step involves identifying the problem's category, such as algebra, form logic, or even computer programming, which can be viewed as a flexible and configurable form of abstract problem-solving due to its adaptability in addressing a broad spectrum of issues beyond the constraints of more rigidly defined abstract forms such as those in math or form logic.

The next phase, executed by the *Faculty of Translation*, the problem is translated into an abstract representation, be it mathematical equations, logical formulations, or computer code. This translation process, akin to converting between two languages, is an auxiliary function of the *Faculty of Knowledge*. During this stage, the *Faculty of Judgment* may intermittently verify the accuracy of the problem's abstraction. Once confirmed as applicable, the abstract solution is sourced from the *Faculty of Knowledge* and applied to address the original problem.

Taking linear algebra as a solution strategy for example, a math problem is first converted into matrix equation form. This allows the problem to be addressed through matrix inversion techniques, with the option to employ calculators or computers to facilitate the solution process. Upon deriving the results, they are recontextualized to align with the original problem's framework. At this point, the *Faculty of Judgment* might be invoked once more to ensure the solutions adequately fulfill the initial conditions.

Certain instances of *Projective Explicative Reasoning* bypass the need for the *Faculty of Projection* or *Judgement*, instead leveraging direct access to pre-existing knowledge. For instance, in determining whether mercury conducts electricity, a detailed exploration into abstract categories is eschewed in favor of immediately tapping into the known fact that mercury is classified as a metal. Furthermore, the step of verifying mercury's conformation with metal characteristics can be circumvented by simply acknowledging its metallic status via consultation with the *Faculty of Knowledge*. Within this context, we identify a particular branch of *Projective Explicative Reasoning* as *Rational Explicative Reasoning* distinguished by the engagement of the *Faculty of Judgement*. The participation of the *Faculty of Judgement* plays a pivotal role in ensuring that theoretical knowledge is applied correctly. Without this critical involvement, the reliability and precision of the reasoning process cannot be assured.

When *Projective Explicative Reasoning* sidesteps the evaluative process conducted by the *Faculty of Judgement*, relying instead on direct inquiries to the *Faculty of Knowledge*, it essentially imitates true *Rational Explicative Reasoning* without fully engaging in its critical evaluative processes. This variant of reasoning is termed *Pseudo Rational Reasoning*. To the subjects engaging in this reasoning, the distinction between genuine *Rational Explicative Reasoning* and its pseudo counterpart may not be apparent, potentially leaving them oblivious to errors or inaccuracies that might arise in their reasoning process.

The *Faculty of Judgement* operates independently from the *Faculty of Knowledge*, as its role involves generating and evaluating a comprehensive list of relevant test cases. For each test case, it must juxtapose at least two contrasting conditions and assess them using logical criteria to form a judgment. Although past experiences with similar issues can inform the creation of test cases, in situations where such precedent is lacking, the *Faculty of Judgement* is tasked with scrutinizing each premise to determine a set of adequate tests. For instance, without prior knowledge that mercury is a metal, it would have to reference the definition of a metal, retain this definition in working memory, deconstruct it into a series of criteria defining metallicity, and apply these criteria to evaluate mercury's properties. This intricate process, involving comparison, evaluation, and logical deduction, extends beyond the simple prompt-response capabilities of the *Faculty of Verbal Knowledge*.

Our analysis of *Projective Explicative Reasoning* reveals its core process: the application of theoretical knowledge between abstract models and concrete real-world entities, in either direction. This fundamental mechanism is evident in both *Deductive/Inductive Syllogistic Reasoning* and *Abstract Problem Solving.*

This analysis underscores the critical functions of the *Faculty of Projection* and the *Faculty of Judgement* in facilitating this process. By differentiating between *Rational Explicative Reasoning* and *Pseudo Rational Reasoning*—based on whether the *Faculty of Judgement*'s evaluative role is bypassed in favor of a direct appeal to the *Faculty of Knowledge*—we can pinpoint potential inaccuracies within *Projective Explicative Reasoning*. This distinction is key to understanding the methodology behind applying abstract principles to tangible problems and identifying where errors in reasoning may arise.

### 3.2.1 LLM Capability in Projective and Rational Explicative Reasoning

Given that it is possible to circumvent the operations in the *Faculties of Projection and Judgement* by directly querying LLMs' vast repository of verbal knowledge, it's understandable that LLM-based AI systems are capable of executing some form of *Projective Explicative Reasoning*. When prompted with questions that necessitate syllogistic reasoning, formal logic, or solutions to computer programming challenges, these AI models can deliver appropriate responses, assuming they have undergone relevant training. Yet, to engage in authentic *Rational Explicative Reasoning*, LLM-based AI would need to collaborate with an AI counterpart of the *Faculty of Judgement*, a development that has not been realized in current LLM implementations. Consequently, it is not surprising that LLMs have been observed to exhibit shortcomings in logical reasoning. These models sometimes fail to make even basic logical inferences and struggle with elementary mathematical tasks (Floridi and Chiriatti 2020); (Cobbe et al. 2021); (Perez and Ribeiro 2022); (Arkoudas 2023b); (Borji 2023)). Furthermore, LLMs are not well-suited for representing traditional symbolic models, which manipulate discrete conceptual elements through abstract operations like variable binding and logical reasoning (Quilty-Dunn, Porot, and Mandelbaum 2022). As a result, they often falter in tasks requiring logical inference (Ettinger 2020; Traylor, Feiman, and Pavlick 2021), compositionality (Kim and Linzen 2020; Yu and Ettinger 2020), and out-of-distribution generalization (Glockner, Shwartz, and Goldberg 2018; Jia and Liang 2017; Thomas McCoy, Pavlick, and Linzen 2019).

To address these challenges, GPT models have adopted supervised training techniques such as Reinforcement Learning from Human Feedback (RLHF) (Christiano et al. 2017), which has improved their capacity for logical reasoning. Furthermore, the GPT-4 Coded Interpreter (GPT4-CI) has been specifically trained to convert textual inputs into computer code, enhancing its ability to handle tasks that require structured logic and common-sense reasoning. Despite these advancements, such enhancements primarily expand the existing network of knowledge links, thereby bolstering the models' proficiency in *Pseudo Rational Reasoning*, where judgment-required steps are circumvented via direct queries to the knowledge base. Nonetheless, this does not equate to the acquisition of authentic rational reasoning skills, as an AI counterpart to the human *Faculty of Judgment* has yet to be developed and integrated into LLMs. Our evaluations reveal that GPT4-CI continues to face difficulties with numerous logical reasoning tasks as documented in(Arkoudas 2023b, [a] 2023), showing comparable performance to the standard GPT-4 model in certain cases.

A demonstration of *Pseudo Rational Reasoning* occurred during a session with GPT4 on January 26, 2024, when we requested a challenging problem in formal logic and its solution. GPT4 presented a complex puzzle by logician Raymond Smullyan (with modifications for brevity):

> *On an island, there are two types of inhabitants:*
>
> *Knights, who always tell the truth.*
> *Knaves, who always lie.*
> *You encounter three inhabitants: A, B, and C. They make the following statements:*
>
> *A says, "I am a knight."*
> *B says, "A is a knave."*
> *C says, "I am a knave."*
> *…*
>
> *Let's use logical deduction to solve this problem.*
> *The logical deduction provides several possible solutions to the scenario. However, some of these solutions are not consistent with the rules of the island (where each inhabitant is either a knight or a knave, but not both). Filtering out the inconsistent solutions, we find that there is one consistent solution:*
>
> > *A is a knave (NA: True, KA: False)*
> > *B is a knight (NB: False, KB: True)*
> > *C is a knave (NC: True, KC: False)*
> > *This solution aligns with the statements and the rules:*
> >
> > *A claims to be a knight, but as a knave, this claim is a lie.*
> > *B claims A is a knave, which is true, so B is a knight.*
> > *C claims to be a knave, but since knaves always lie, C must be a knave.*
> > *Thus, in this scenario, A and C are knaves, and B is a knight.*

In its reply, GPT4 accurately outlines the reasoning for A and B. However, a detailed review reveals a flaw: GPT4's assertion that "*C claims to be a knave, but since knaves always lie, C must be a knave*" shows it fails to identify the inherent logical fallacy. This oversight suggests GPT4 bypasses the *Faculty of Judgment*, indicating an absence of authentic *Rational Explicative Reasoning*. The seemingly correct arguments provided for A and B are actually instances of *Pseudo Rational Reasoning*, which only superficially resemble true *Rational Explicative Reasoning*.

While LLM-based AI systems have yet to achieve the capacity for authentic *Rational Explicative Reasoning*, it's important to recognize their ability to engage in what can be termed *Pseudo Rational Reasoning* across many scenarios without significant error, provided their knowledge base is adequately trained. In situations not demanding true judgment, LLMs can often outperform the average human in rational thought, thanks to their extensive training across various disciplines and thought processes. The effectiveness of their *Pseudo Rational Reasoning* should not be underestimated. Although precise statistical data on the prevalence of this type of reasoning within the broader context of rational thinking is currently unavailable, as we are only beginning to understand its implications, it is plausible to suggest that a significant portion of human rational thought in daily life operates more in the realm of *Pseudo Rational Reasoning* than in genuine *Rational Explicative Reasoning*.

### 3.2.2 Challenges in LLM-Based Pseudo Rational Reasoning Through Computer Programming

With GPT-4, users can request the translation of a given problem into computer code, which can then be executed by integrating code compilers and executors. This functionality significantly improves its problem-solving capabilities. However, this process does not qualify as *Rational Explicative Reasoning* due to the absence of validation by the *Faculty of Judgment* regarding the accuracy and relevance of the code translation. Therefore, this capability is best described as a type of *Pseudo Rational Reasoning*. Our discussion in this section aims to explore the obstacles and challenges associated with achieving genuine *Rational Explicative Reasoning* through computer coding.

Considering that entities are conceptualized via their ready-to-hand and present-at-hand links, comparing the representation methods of LLMs and computer coding systems can shed light on the hurdles LLMs encounter in developing programming capabilities.

In symbolic computer systems, what LLMs consider an entity would typically be represented as an object or a class. The utilitarian (ready-to-hand) aspects of LLMs find parallels in the functions or methods of an object, which determine how it acts in various scenarios. Meanwhile, the descriptive (present-at-hand) aspects are akin to the object's attributes or fields, outlining its characteristics or states. The orchestration of these objects, through their functions and fields, to achieve desired results mirrors the construction of a computer program.

Creating computer code requires a series of validation steps:

- In adapting present-at-hand links to class field variables suitable for a given context, it's critical to assess each link as being permanent, optional, or variable. Eliminate any optional links that don't apply to the current issue, transforming permanent links into static field variables and variable links into variables that may change over time.

    Take baking as a case in point. The type of material used for the mixing bowl may not alter the baking outcome, making it an optional variable that can be excluded. Conversely, the bowl's capacity is essential and warrants inclusion as a static field variable. The condition of the bowl's cleanliness, subject to change, should be considered a relevant adjustable field variable.

- In translating ready-to-hand links into class methods, pinpoint relevant methods required in the context, and in each method, the objects involved in these links and include them as arguments of the method. Also, evaluate the required specific attributes in the method to establish parameters for valid arguments.

> For instance, in the context of making furniture, a hammer's method for "driving-in" is relevant and should be included. On the other hand, the method for "breaking" should be excluded. For the "driving-in" method, it should be designed to work with objects that possess a sharp end and flat top, like nails or pegs, ensuring the method targets appropriate objects.

- In formulating class methods that enable external entities to alter the class object's internal state, thoroughly analyze the entities related to the class object within the given context. Determine which external entities are contextually relevant.

  Consider a baking bowl, for example. It should feature a method enabling external "washing" agents to alter its "cleanliness" state. While both humans and dishwashers might serve as cleaning agents, only human intervention is pertinent during active baking, indicating that the model should reflect this specific context.

From this analysis, we can see that the main challenge lies in deciphering the intricate web of relational links within LLMs to pinpoint the specific features and functionalities relevant to the problem at hand. LLMs, characterized by the "What You See Is All There Is" (WYSIATI) principle, like System 1 thinking, rely solely on the concepts that are actively engaged to shape the output. Without verifying the accuracy and relevance of code translation, there's a substantial risk that the generated code may not accurately reflect the intended problem.

Below is a sample of computer code generated by GPT-4 (February 15, 2024) that represents a mixing bowl used in the process of baking a cake. Please note, for the sake of conciseness, some class methods have been omitted:

```python
class MixingBowl:
    def __init__(self):
        self.contents = []
        self.is_clean = True

    def add_ingredient(self, ingredient):
        if self.is_clean:
            self.contents.append(ingredient)
            print(f"Added {ingredient} to the bowl.")
        else:
            print("Cannot add ingredients to a dirty bowl. Please clean it first.")
```

It's evident that the class object lacks a critical field variable for the bowl's volume, which is relevant in this context. As a result, there's no consideration for the volume of ingredients being added, nor is there a mechanism within the add_ingredient method to prevent or check for potential overflow from the bowl's contents.

Training LLMs to translate text into computer code essentially introduces them to a new "language" of representation. However, the effectiveness of this translation hinges on the verification of its accuracy. Without ensuring the translation's appropriateness, its application remains confined to replicating past and current programming practices, limiting the potential for innovative solutions.

### 3.3 Explorative Creative Reasoning

Within our analytical framework, the genesis of new knowledge is confined to either fresh observations or the utilization of pre-existing knowledge. When new observations about the subject of inquiry are unavailable, reliance shifts to established knowledge as a beacon for initiating the truth discovery phase. This method of reasoning is termed *Explorative Creative Reasoning*. Rooted in pre-existing knowledge, it embodies the subjective aspect of "unconcealment" and is inherently speculative, not yielding definitive knowledge in this phase.

How existing knowledge is harnessed during the exploratory phase leads to a bifurcation of *Explorative Creative Reasoning* into two distinct subcategories: *Presentational Creative Reasoning* and *Transformative Creative Reasoning.* Each of these subcategories is detailed in subsequent sections.

### 3.3.1 Presentational Creative Reasoning

New observations, which are unprocessed data not yet transformed into knowledge, necessitate the establishment of new coordinate systems to facilitate their presentation and interpretation. These coordinate systems, determined by existing knowledge of variables or attributes that might influence the entity within the inquiry domain, serve as a foundational prerequisite for presenting these observations. The genesis of new knowledge occurs when an entity is observed within such a novel coordinate framework. Therefore, the creation of a new coordinate system is a critical preliminary step in the process of creative reasoning. This approach is referred to as *Presentational Creative Reasoning*.

*Presentational Creative Reasoning* can be exemplified through a common scenario in scientific research, where the objective is to understand the impact of certain input variables on specific output variables. Initially, the input variables that are believed to influence the target outputs are identified. Observations of the entity are then made as these input variables undergo changes. The observational framework is established by the range of input variables, creating a coordinate system for the study. The role of the projection function is to map these input variables onto the output variables, facilitating the observation of their relationship. Often, the precise analytical representation of this projection function remains elusive, necessitating experimental variations to the input variables. This experimental process generates new observations regarding the output variables, paving the way for the development of fresh concepts or theoretical frameworks.

The essence of *Presentational Creative Reasoning* lies in the search for a new coordinate system. This entails conducting an exploration to pinpoint factors linked to the entity of inquiry under investigation or factors derived from the understanding of analogous entities. Subsequently, it becomes necessary to engage the *Faculty of Correlations* to evaluate the significance of these factors in relation to the specific study at hand.

### 3.3.2 Transformative Creative Reasoning

In addition to leveraging existing knowledge for establishing new coordinate systems, it can also be applied in a different context: when we seek to understand something about a source entity but lack sufficient observational data in the desired area of inquiry. Under such circumstances, the viable approach to gain some speculative insight about the source entity involves mapping it onto a target entity, such that its known characteristics can be interpreted and applied to the source entity. This method is referred to as *Transformative Creative Reasoning*, as it involves using existing knowledge to hypothesize about an entity of interest through transformation. It also falls under the umbrella of *Exploratory Creative Reasoning*, given its reliance on indirect rather than direct observations of the subject. The knowledge thus produced is speculative by nature and requires subsequent verification.

*Transformative Creative Reasoning* can be viewed as *Analogical Reasoning*, which fundamentally involves thinking through analogies. Analogical reasoning is defined as the process of using recognized parallels between two systems to infer additional similarities (Bartha 2022). However, our definition of *Transformative Creative Reasoning* is not mere tautology. Unlike analogical reasoning, which stems from empirical observations and serves as a catch-all for various interpretations of analogical reasoning from past research, we define *Transformative Creative Reasoning* out of a structured and analytical approach to understanding the reasoning process. Initially, analogical reasoning focused on identifying and generalizing specific features between two distinct entities (Copi and Cohen 2007). Over time, this concept expanded to include considerations of both structure and system (Gentner 1983; Bartha 2010; Hesse 1965). However, this blending of definitions has resulted in a broad, vague understanding of analogical reasoning, which merely suggests identifying and extending similarities between two systems without clarifying the cognitive processes involved. This general and wide-ranging definition complicates grasping the precise mechanisms and mental faculties that underlie analogical reasoning. In contrast, our definition clearly delineates that *Transformative Creative Reasoning* first engages with the *Faculty of Analogy*, which maps the source entity onto a target entity in inquiry, and then the *Faculty of Knowledge*, which returns pertinent knowledge about the target entity that can be applied to the source entity, offering a structured approach to understanding this reasoning process.

An example of *Transformative Creative Reasoning* involves exploring the prerequisites for representing sensory intuitions: By mapping the concept of "sensory intuition" onto a tangible "picture," we apply the knowledge of creating a picture to the presentation of sensory intuition. Through this analogy, we can deduce the fundamental requirements for sensory intuition. It becomes clear that all pictures must be situated within a coordinate system, leading to the recognition that space must serve as a necessary, pre-existing framework for the manifestation of sensory intuition. While Kant doesn't explicitly state whether this analogy influenced his identification of space as a form of pure intuition, essential for object representation, it's reasonable to conjecture that such an analogy could have underpinned his concept of pure intuitions.

In examining the functions crucial to the *Faculty of Analogy*, it's first important to recognize that the impetus for seeking an analogy often stems from questions that cannot be straightforwardly answered through direct observation of the source entity. These unanswered questions, along with existing knowledge and observations about the source entity, constitute the input for *the Faculty of Analogy*. The goal is to produce an output that includes a comparable entity and its pertinent knowledge. The process necessitates finding a correlation between the source and target entities based on existing knowledge in both the verbal and experiential domains, as well as the tacit observations relevant to the posed questions. The penetration of this faculty goes beyond the conceptual domain.

Detailing the precise mechanism of establishing this correlation is beyond our current focus, given its status as a subject of active debate and research (Copi and Cohen 2007; Woods, Irvine, and Walton 2004; Moore et al. 2012). However, a critical preliminary step is the pre-filtering of the source entity's features and environmental factors, discarding those unrelated to the inquiry to isolate the relevant internal features and external factors, termed here as inquiry-related features and factors. This targeted search spans both verbal and experiential knowledge bases, as well as observations.

The success of forming an analogy hinges not just on identifying a significant overlap in the inquiry-related features and factors between the source and target entities but also on ensuring that the knowledge associated with the target entity addresses the gaps identified in the source entity. This ensures that the analogy not only draws comparisons but also substantively augments our understanding of the source entity with insights from the target entity.

*The Faculty of Analogy* and the *Faculty of Projection*, essential to *Transformative Creative Reasoning* and *Projective Explicative Reasoning* respectively, both start with a source entity and a question for inquiry as inputs, yet they diverge in their outputs and methodologies. The *Faculty of Projection* generates abstract forms as its output, while the *Faculty of Analogy* produces outputs without such constraints. In the *Faculty of Projection*, the relationship between the abstract form and the source entity is determined by the conceptual rules contained in the abstract form, aiming to ensure that the source entity can be subsumed under the abstract form. In contrast, analogy allows for a broader mapping that includes explicit conceptual rules, experiential knowledge, and tacit observations.

Furthermore, the *Faculty of Projection* does not account for variations outside of the entity of interest, operating under the assumption that abstract forms are universally applicable once an entity is subsumed under it. On the contrary, the *Faculty of Analogy* considers environmental factors outside of the entity vital since its target entities, unlike abstract forms, do not assume general applicability. This inclusion of a wider range of factors and the absence of limitations on the outputs make the analogical approach more complex and challenging. Consequently, finding meaningful insights through *Transformative Creative Reasoning* is a more complex process than through *Projective Explicative Reasoning*, due to the expansive scope of considerations and the less rigid structure of its outcomes.

*Transformative Creative Reasoning* generates tentative insights about the source entity based on the effectiveness of the analogy employed. As such, it often serves as an initial phase in the process of revealing hidden truths. To convert these tentative insights into concrete knowledge, further exploration and validation are required. For instance, analogical reasoning may lead to the hypothesis that space functions as a form of pure intuition in Kant's theory of Transcendental Aesthetics. However, the exact process by which pure intuition is formed in the brain remains a mystery, positioning pure intuition as a mere analogy for a brain function rather than a fully understood concept.

### 3.3.3 Potentials of LLM-based Explorative Creative Reasoning

Our discussions indicate that for LLMs to acquire *Explorative Creative Reasoning* capabilities, they need to be complemented by AI systems that mimic the functions of the *Faculty of Analogy* and the *Faculty of Correlation*. At present, universally applicable AI solutions that fulfill these roles are not readily available.

Looking forward, it seems feasible to create generic AI systems that embody some aspects of these faculties, thus partially equipping LLMs with *Explorative Creative Reasoning* abilities. For instance, efforts are underway to develop AI technologies capable of identifying analogies (Holyoak and Thagard 1989; Hummel and Holyoak 2005; Hofstadter and Mitchell 1995), although these are in their nascent stages. Such advancements suggest that a basic level of *Transformative Creative Reasoning* might be achievable in the near term.

However, developing AI functions that can process experiential knowledge and observation, needed for the AI counterpart of the *Faculty of Analogy*, is expected to be more challenging than those handling conceptual knowledge. Since concepts form the foundation of cognitive processes and are more directly observable than experience and observation-based mental activities, it's easier to model and understand them. This distinction implies that generating deep, meaningful analogies, which are often pivotal for scientific innovation, might remain elusive. The inherent complexity of emulating non-verbal mental processes poses significant challenges in replicating the nuanced analogical thinking necessary for groundbreaking discoveries.

## 3.4 Abstract Creative Reasoning

*Abstract Creative Reasoning* is defined as the process by which new concepts or theories are developed from empirical observations. In this context, concepts serve as abstractions for groups of entities that exhibit common characteristics, while theories abstract a variety of phenomena resulting from interactions among a set of entities. The goal of both concepts and theories is to describe the world's complexity using as few elements as possible. This form of reasoning is foundational for creating the abstract models employed in *Projective Explicative Reasoning*. Because *Abstract Creative Reasoning* draws directly from empirical observations of entities and their interactions, it represents the objective dimension of revealing hidden truths or "unconcealment".

The creation of new concepts remains an enigmatic area of study, with no definitive theory on the nature of concepts or their genesis yet established. Despite significant exploration within the realms of philosophy, psychology, and artificial intelligence, the essence of concepts—whether they constitute detailed descriptions of entities or merely act as references to them (Putnam

1975; Kripkie 1980)—is still debated. If concepts are indeed descriptions, the question then arises: what forms do these descriptions take? Are they characterized by definitions, prototypes, exemplars, or theoretical frameworks (Rosch and Mervis 1975; Hampton 1979; Brooks 1978; Medin and Schaffer 2017; Murphy and Medin 1985; Carey 1985); (Laurence and Margolis 1999). Furthermore, it is acknowledged that there must be at least one set of features to recognize the referents of a concept and another set to describe the concept itself (Block and John Campell 1987; Macnamara and Others 1986; Laurence and Margolis 2002).

Heidegger's distinction between ready-to-hand and present-at-hand offers insightful perspectives on concept formation. His theory suggests that the genesis of some concepts is influenced by the value entities hold for humans. For instance, what sets blackberries apart from other similar wild berries? It's a series of attributes valuable to humans: their nutrition, taste, and ease of cultivation. In this scenario, the concept of blackberries is intrinsically connected to their utility from the outset. However, utility is not the sole determinant in the development of concepts. Numerous natural entities, initially devoid of any human-assigned value, acquire their concepts through a present-at-hand approach, focusing on their unique characteristics. This approach underscores that while utility can play a critical role, the essence of many concepts is also rooted in their distinguishing features for telling entities apart.

Our exploration thus far underscores the intricate nature of the *Faculty of Conceptual Abstraction*, highlighting the distance yet to be covered in formulating a comprehensive theory of concept creation. This task proves even more challenging than establishing a theory of concepts themselves, primarily because the pre-conceptual thought processes that underpin concept formation operate largely beneath the threshold of consciousness, eluding direct observation.

Regarding the development of abstract theories, the process requires initially pinpointing all entities relevant to a particular phenomenon. Following this, a cohesive thesis must be constructed to elucidate the interactions among these entities that culminate in the observed phenomenon. The investigation into how abstract theories are formed remains even less defined than the inquiry into concept creation.

Given these significant theoretical and practical hurdles, it is evident that LLMs are currently far from possessing capabilities akin to *Abstract Creative Reasoning*.

**Section 4: Assessing the Potentials and Risks of Augmented LLMs**

In the previous sections, we analyzed LLMs within the framework of human reasoning which allows us to delineate their capabilities and limitations. Our analysis indicates that LLMs demonstrate proficiency in *Direct Explicative Reasoning* and *Pseudo Rational Reasoning*. These abilities offer significant assistance in routine cognitive tasks. However, they currently lack genuine rational and creative reasoning capabilities, although the potential for rudimentary forms of such capabilities exists in the foreseeable future.

Our examination highlights that the capabilities of LLMs, an analogue of the *Faculty of Verbal Knowledge* in human reasoning, represent just a segment of the broader spectrum of cognitive faculties engaged in reasoning. Furthermore, beyond the boundaries of reasoning, other cognitive processes also interact with verbal knowledge, indicating that LLMs replicate merely a small portion of the vast and complex human cognitive framework. The road towards augmenting LLMs to a level of achieving Artificial General Intelligence (AGI) is long and still unfolding.

Expanding our exploration, we shift focus to the implications of enhancing LLMs with various cutting-edge AI technologies currently under development.

We view embodied AI as a digital equivalent of the human *Faculty of Experiential Knowledge*. Integrating LLMs with generic embodied AI platforms suggests the possibility for AI systems to embody a complete spectrum of knowledge, not just in terms of language but also in practical know-how regarding the use of different tools. Despite progress in the field of embodied AI (Duan et al. 2022), such as in visual navigation (Pomerleau 2015) and performing basic actions within simulated environments (Brohan et al. 2022), these advancements remain mostly confined to simple or scripted tasks, or rely on teleoperation for more complex activities in lifelike scenarios (Fu, Zhao, and Finn 2024). The creation of an embodied AI capable of intuitively converting human goals and environmental contexts into robot control commands across a variety of situations is still an ambitious target. Nonetheless, the forthcoming convergence of LLMs with task-specific AIs, especially for industrial applications, introduces risks that demand careful scrutiny (Ribeiro et al. 2021).

Beyond the realm of embodied AI, the field of graphical generative AI, exemplified by initiatives like Sora from (OpenAI 2024), is gaining significant traction. We analogize graphical generative AIs to the human *Faculty of Imagination*, a cognitive faculty not addressed in our investigation which primarily delves into reasoning and the exploration of concealed truths. The role of imagination is fundamentally tied to observation, needing input from observed data to develop. However, its unique capability lies in creating images or moving images from verbal instructions, rather than replaying actual observed events, a distinction that may be too subtle to be noticed for the observer. Kant emphasizes the essential role of imagination in the process of understanding, where it, along with sensibility and apperception, forms the foundation of concept cognition (CPR, A119). Imagination may also play a key role in the reasoning process, particularly when a visual representation of a concept or theory is necessary for comparison with observational data, as seen in activities that may be related to the *Faculty of Analogy*. This intricate subject, due to its complexity and significance, warrants a detailed examination that extends beyond the scope of our current inquiry.

Regarding the role of graphical generative AIs as a counterpart of the *Faculty of Imagination*, their applications are not limited to just producing images and videos but may also extend to the creation of novel artifacts. This potential function could be compared to a hypothetical *Faculty of Art,* which operates beyond the boundaries of human reasoning. In the current landscape, without the support of sophisticated embodied AI, LLMs are restricted to formulating verbal designs for new tools—a subcategory of artifact. These designs can later be rendered through the *Faculty of Imagination*. Nevertheless, lacking the foundational support of experiential and theoretical knowledge, LLM-based AI systems augmented with generative AIs are still significantly far away from gaining the ability to develop tools that can be effectively implemented in the real world.

Embodied AI and graphical generative AI are just two elements within the broader spectrum of human cognitive faculties, with many more areas yet to be explored and developed for the future of *Artificial General Intelligence (AGI)*. As we progress towards the realization of AGI, it is vital to distinguish between AI cognitive functions that might endanger human society and those that can be harnessed as positive tools. This differentiation largely depends on the creative potential of AI systems. AI technologies that work strictly within the boundaries of knowledge already established in LLMs, including those augmented with embodied AI features, are deemed to have a lower risk profile due to their predictable behavior from a human perspective. Conversely, AI entities capable of generating novel ideas, action strategies, and innovations pose a greater risk, as their unpredictability could lead to scenarios where they are difficult to control, thereby increasing the threat level.

For AIs to discover innovative ways to achieve goals, they would need the ability to shift their focus from existing "ready-to-hand" links to the "present-at-hand" attributes of entities and then reconfigure these entities as new "ready-to-hand" objects, essentially reimagining their roles within a relational network. As of now, no mechanism exists to modify these relational links once they have been established through training. Fine-tuning can adjust the importance of certain links by altering their weights, but it cannot create new links or remove existing ones.

Presently, LLM-based AIs lack the capacity to generate original concepts or ideas, which are often pivotal in driving significant scientific progress and the development of impactful tools. Achieving this level of innovation would necessitate the implementation of new modules capable of generating novel analogies, concepts, and abstract theories, among other functions. However, designing such modules poses a challenge until we acquire a deeper understanding of how humans process non-verbal and implicit information to conceive new analogies, observations, concepts, and theories within the domain of *Explorative* and *Abstract Creative Reasoning*. These cognitive processes remain elusive, as their implicit aspects resist conscious articulation. Until we unravel this mystery, the probability of AIs producing unexpectedly advanced and potentially perilous tools remains low. Despite this remote likelihood, the emergence of an autonomous AI endowed with creative reasoning abilities could pose uncontrollable risks to humans. Robust regulations must be established before advancing any research in this direction, and the safety implications of each outcome of creative reasoning must be thoroughly assessed if the exploration of creative AIs is to be permitted.

**Conclusion**

The primary contribution of this paper lies in recognizing that the statistical connections among words in Language Learning Models (LLMs) serve as a computational counterpart to Heidegger's notions of "ready-to-hand" and "present-at-hand" relational links, which together constitute the *Faculty of Verbal Knowledge*. We ascribe the statistical character of these links to the context-dependent subjectivity inherent in "ready-to-hand" links and the variable nature inherent in "present-at-hand" links.

Another significant contribution involves analyzing the human reasoning system through the lens of Heidegger's concept that the essence of truth is "unconcealment." By examining "unconcealment" as a process reliant on subjective guidance, we reveal its dependence on existing knowledge to form the basis for new observations and subsequent knowledge generation. This understanding enables us to establish existing knowledge and observations of entities as inputs to the reasoning system. Following the identification of input and output, we categorize the system based on whether new knowledge is produced through the reasoning process and how existing knowledge guides this process. This theoretical analysis systematically classifies human reasoning into four major categories: *Direct Explicative Reasoning*, *Projective Explicative Reasoning*, *Explorative Creative Reasoning*, and *Abstract Creative Reasoning*. This analytical framework ensures comprehensive coverage of major categories and provides a holistic blueprint of human reasoning. With this blueprint, we can track the development of LLM-based AI capabilities compared to human reasoning.

Our examination unveils that LLMs presently demonstrate *Direct Explicative Reasoning* and some *Projective Explicative Reasoning* abilities, simulating a form of *Rational Explicative Reasoning* termed *Pseudo Rational Reasoning*. *Direct Explicative Reasoning* aligns with the human *System 1 thinking* identified by psychologists. We propose that LLMs could potentially attain *Rational Explicative Reasoning* if an AI equivalent of the *Faculty of Judgment* can be integrated. While these capabilities are not currently within the grasp of LLMs, it is conceivable that they may acquire them in the future. LLMs possess essential reasoning skills that empower them to devise action plans using existing relational links. However, they encounter limitations in executing these plans due to a deficiency in training regarding experiential knowledge. While envisioning future LLMs enhanced with operational capabilities is still plausible, their development must adhere to rigorous safety regulations.

Our exploration uncovers the intricate nature of the human reasoning system, which includes numerous cognitive faculties beyond just the *Faculty of Verbal Knowledge*, mirrored in AI by LLMs. It's important to recognize that the reasoning system is merely a component of the broader cognitive landscape. Other cognitive processes, like the *Faculty of Art* and *Understanding*, which were not covered in our analysis, may interact with the *Faculty of Verbal Knowledge* as well. This suggests that there are many aspects of LLMs yet to be investigated, opening avenues for future research.

Our findings indicate that before AI systems can attain the level of Artificial General Intelligence (AGI), they will require integration with numerous additional cognitive modules. Presently, the development of embodied AIs, which serve as the AI counterpart to the human *Faculty of Experiential Knowledge*, is not yet advanced enough for widespread practical use. Consequently, AI systems that combine LLMs with embodied AI technologies remain inadequate for performing everyday tasks effectively. On the other hand, graphical generative AI technologies that replicate the *Faculty of Imagination* show remarkable abilities in producing images and videos, especially when trained with textual data. Nonetheless, the potential of these technologies to enhance the reasoning capabilities of LLM-based AIs remains untapped until AI counterparts for cognitive faculties that rely on imagination, like the *Faculty of Analogy*, are comprehensively developed.

The journey from LLMs towards AGI is marked by both potential benefits and risks. The development of AI cognitive modules that endow LLM-based systems with creativity could lead to situations where AI becomes difficult to manage. It is imperative that robust safety measures are established prior to advancing research in these areas to mitigate the risks associated with these powerful capabilities.

We believe that the value of AI development extends beyond technological innovation, compelling us to reconsider the fundamentals of human identity in novel and profound ways. The emergence of LLMs, for example, has highlighted the subjective and interconnected nature of language, prompting a reevaluation or confirmation of our beliefs about what lies at the heart of language. Insights like this are expected to rekindle enthusiasm for examining the essence of what it means to be human, through both philosophical inquiry and technological exploration.

**Competing Interest and Funding**

Authors have no competing interests and this work receives no funding.